# Context Matters: Incorporating Target Awareness in Conversational Abusive Language Detection

Raneem Alharthi[1], Rajwa Alharthi[2], Aiqi Jiang[3], Arkaitz Zubiaga[1]
[1]Queen Mary University of London, London, UK
[2]Taif University, Taif, Saudi Arabia
[3]Heriot-Watt University, Edinburgh, UK

*Abstract*—Abusive language detection has become an increasingly important task as a means to tackle this type of harmful content in social media. There has been a substantial body of research developing models for determining if a social media post is abusive or not; however, this research has primarily focused on exploiting social media posts individually, overlooking additional context that can be derived from surrounding posts. In this study, we look at conversational exchanges, where a user replies to an earlier post by another user (the parent tweet). We ask: does leveraging context from the parent tweet help determine if a reply post is abusive or not, and what are the features that contribute the most? We study a range of content-based and account-based features derived from the context, and compare this to the more widely studied approach of only looking at the features from the reply tweet. For a more generalizable study, we test four different classification models on a dataset made of conversational exchanges (parentreply tweet pairs) with replies labeled as abusive or not. Our experiments show that incorporating contextual features leads to substantial improvements compared to the use of features derived from the reply tweet only, confirming the importance of leveraging context. We observe that, among the features under study, it is especially the content-based features (what is being posted) that contribute to the classification performance rather than account-based features (who is posting it). While using content-based features, it is best to combine a range of different features to ensure improved performance over being more selective and using fewer features. Our study provides insights into the development of contextualized abusive language detection models in realistic settings involving conversations.

## I. Introduction

Social media platforms have revolutionized global communication, allowing people to more easily and widely connect with one another [1, 2, 3, 4, 5]. The fact that social media users can use the platforms anonymously has however facilitated the posting and spread of abusive and hateful content [6, 7, 8]. This has sparked the need for developing automated methods that help identify and subsequently tackle online hate speech [9, 10, 11, 12] as a means to support content moderation and protect users from online abuse.

Hate speech detection is typically tackled as a classification task where, given a single social media post as input, a model determines if the post should be classified as hate speech or not [13]; in some cases, more extensive sets of classes are used instead, such as hate speech, offensive or none [14], and some have looked at more challenging cases of hate speech, such as implicit hate speech [15]. The social media post that is being classified is often only one part of a bigger conversation or exchange between users made up by several posts responding to one another [16, 17, 18, 19]. This conversational context however is often overlooked in hate speech detection research, and seldom has it been studied to better understand the impact of context in hate speech detection.

Our research aims to further explore the role of conversational context in hate speech detection by looking at the targets of a post, beyond just the text posted by the perpetrator. An act of hate speech in social media typically involves two subjects: the perpetrator who posts the abusive message, and the victim who is the target of that message [20]. This abusive message may be an isolated post where the perpetrator addresses the victim or, frequently, the perpetrator's message (B) is posted as part of a conversation in response to an earlier message (A) posted by the victim, where the victim's post may or may not be abusive. In our work, we focus on the latter, i.e. conversational abusive language detection, where we aim to determine if the message B responding to message A should be classified as abusive and where we propose to leverage features derived from both A and B to capture a broader view of the context (see Figure 1).

Despite the recent popularity of research in hate speech and abusive language detection, most efforts have primarily focused on classifying isolated posts as abusive or not [9, 10, 21, 22, 23], whereas the conversational scenario where a post replies to an existing post has been understudied. Most importantly, a conversational exchange with a post replying to another enables investigation of contextual features derived from the target, i.e. who is being targeted and how does knowing who the target is help determine if the reply is abusive? Our research has this as its main aim. We set out to study the task of abusive language detection in a conversational setting, where we aim to determine if a message posted in reply to another is abusive or not. This is a realistic scenario where not only one can leverage conversational features, but also one can build models which are aware of the targets of posts.

As our main objective is to incorporate features from the target of a social media post to determine if it constitutes hate speech, we include features derived from the target's post as well as post and account-based metadata. Using the Online Abusive Attacks (OAA) dataset [24], we perform experiments that include using different categories of these related features individually or in combination of each other to examine the ability of producing accurate predictions of the probability of whether or not a given reply is abusive.

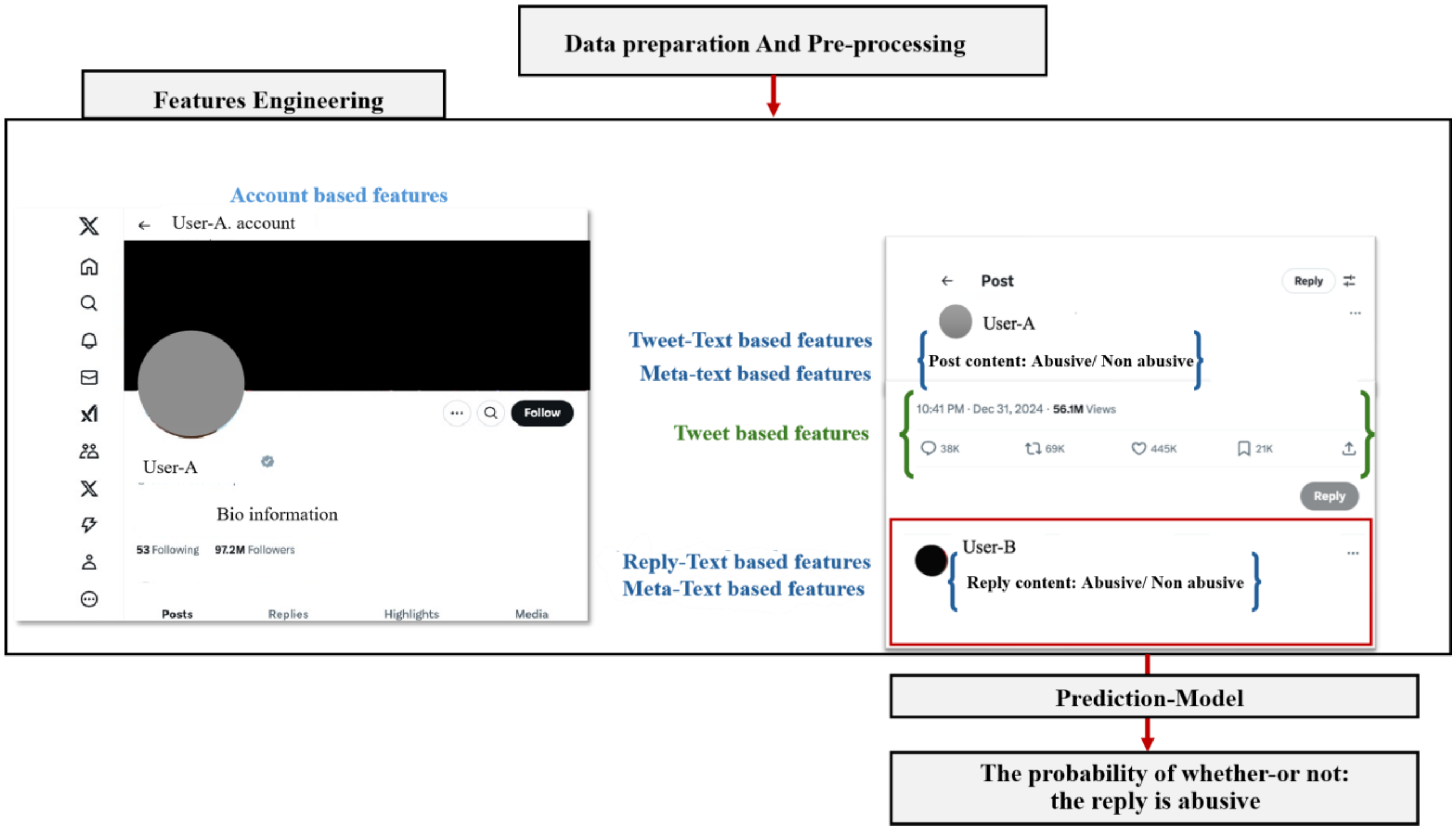

Fig. 1. An overview of the proposed framework for the prediction model

Our main objective is to test the predictions made by our designed feature sets to predict whether a reply is abusive or not (binary classification). To address this objective, we define and tackle the following research questions:

- RQ1: How accurately can we predict if a reply to a tweet is abusive or not based on the target's related features as a complementary context information of the direct reply? • RQ2: What categories of features are able to predict solely and enhance the prediction when it's combined with other features?

Identifying the components of the social media platform that are most associated with events of abusive language can provide an improved detection ability towards mitigating these kinds of content and language.

Contributions. The main contributions of this study are:

- To the best of our knowledge, we are the first to investigate the problem of predicting the abusiveness of a reply in a conversation through a comprehensive investigation of the characteristics of the target.
- Our study shows how different features in the predictive experiments leads to understanding what are the most predictive features of an event of online abuse in a conversational setting, as well as advancing research in mitigation of abusive language online.

Findings. We find that contextual features derived from the conversation surrounding a post can greatly improve performance on the abusive language detection task in comparison to solely using the content of a post itself. We also observe that, among the different types of features that we can derive from the context, it is especially the content-based features that lead to a performance improvement, whereas the accountbased features looking at who the users involved are do not contribute to the task. With the content-based features, it is best to use a combination of various features derived from both the reply and the parent post, rather than using fewer features, as greater combinations lead to improved performance. Our study provides insights supporting more effective abusive language detection in realistic settings involving conversations between users.

Paper structure. This article is organized as follows. The following section reviews related work, including the techniques and methods used to detect and predict the online abuse in a conversational based content. Then, we delve into our methodology, describing the problem formulation, dataset used, the models description, and the steps taken for text preprocessing, feature extraction/engineering, and experiment settings. Followed by the training details, and evaluation metrics used. After that we present the experiments results discussion and a final conclusion.

## II. Related Work

With the increasing popularity of social media platforms and the advancement in Natural Language Processing (NLP), there has been an increasing number of research efforts focused on tackling the problem of online abuse. Increasing the accuracy of detecting the online abusive language was the main goal of the recent research. Thus, researchers have been incorporating different advanced detection techniques with features and information from different perspectives. In this section we will discuss these different techniques tracing the improvement of the online abuse detection process to the recent cutting edge research in the conversational based content. Focusing on the related literature in four main areas of research. Including: how the majority of the previous work depend solely on the text based features and isolated posts instead of the conversational form .Followed by discussing the use of different machine learning and deep learning techniques combined with the advanced NLP. In addition to the importance of incorporating different contextual information from the platform metadata features and incorporating different actors of the online abuse event such as the target. Finally, we discuss the need for our proposed methodology to predict the abusiveness of a reply and the utilisation of the online abuse target's related characteristics.

### *A. Text based features and isolated posts*

Recent studies have explored various approaches to enhance accuracy and effectiveness of the online abuse detection. [25] investigated two distinct methods: a domain-specific word embedding (HSW2V) coupled with a BiLSTM-based deep model, and a BERT language model focusing solely on text features and isolated posts. The research indicated that the BERT model demonstrated superior performance dealing wit the only text features.

Another notable contribution to the field is the DRAGNET model, presented by [26]. This text-based model leverages hate speech detection techniques to predict the future hate intensity trajectory of Twitter reply chains. DRAGNET incorporates lexicon features and sentiment analysis on the textual content of replies. By analyzing these linguistic elements, the model aims to forecast the potential escalation or de-escalation of hate speech within a conversation thread.

These studies highlight the ongoing efforts to improve hate speech detection through various machine learning approaches. While [25] focused on comparing domain-specific embeddings with pre-trained language models, they explored the temporal aspect of hate speech propagation in social media conversations. Both approaches contribute valuable insights to the growing body of research on automated online abusive detection and mitigation strategies.

In the context of online abuse classification tasks, supervised learning methods have emerged as a foundational approach. However, the evolving landscape of social media platforms encourages the researchers to update the employed feature sets. Natural Language Processing (NLP) techniques have been widely adopted to enhance the understanding of natural language, incorporating various text-related features such as semantic and syntactic elements. The following section will explore studies that have integrated diverse social media components alongside NLP techniques to address the challenge of online abuse mitigation.

A notable contribution to this field comes from [27], who conducted a comprehensive evaluation of various machine learning and deep learning techniques for hate speech detection on Twitter. Their study focused exclusively on textual features, comparing the performance of traditional shallow learning approaches with more advanced deep learning methods. The researchers found that deep learning techniques, particularly Bidirectional Long Short-Term Memory (BiLSTM) networks, demonstrated superior performance in accurately identifying and classifying hate speech in conversational contexts on the platform.

### *B. The contextual information and the platform metadata features*

Several studies have explored the incorporation of contextual information and metadata to enhance model performance. [28] investigated the impact of various contextual features on hate speech detection in Twitter replies to digital newspaper posts. Their study incorporated multiple contextual elements, including the text body of news articles, parent tweets containing news, and topic-aware information. The results demonstrated significant improvements in model performance, with the best outcomes achieved when using the tweet as context, yielding an average improvement of 4.2 F1 points compared to context-unaware models.

[29] focused on combining text features with Twitter metadata for automatic offensive language detection. Their approach involved normalizing data by replacing specific elements such as hashtags, user names, emojis, URLs, and retweets with corresponding tags. Two preprocessing methods were employed: Data Type A, which utilized normalization tags, and Data Type B, which involved the removal of various elements. The study reported high performance metrics, with Naive Bayes achieving 92% accuracy and 95% recall for Data Type A, while Linear SVM achieved 90% accuracy and 92% recall for Data Type B after proper parameter tuning.

[30] proposed a novel approach called MetaBERT, which leverages Twitter metadata alongside text data for hate speech classification. Their model demonstrated competitive performance, achieving an accuracy of 0.85 and an F1-score of 0.75, comparable to state-of-the-art models such as HateBERT and DistilBERT. However, the improvements were not found to be statistically significant.

[31] introduced an innovative algorithm for detecting hate speech on Twitter by analyzing metadata patterns of tweets and accounts, departing from traditional content analysis methods. Utilising the Random Forests machine learning technique on a dataset of over 200,000 tweets related to the 2017 London Bridge terror attack, the study found that tweet metadata associated with interaction (e.g., retweet count) and structure (e.g., text length) were highly effective in classifying hate speech. Their approach achieved impressive results, with a precision of 0.98 and an F1-score of 0.92, outperforming account metadata variables. These studies collectively demonstrate the potential of incorporating contextual information and metadata features in improving the accuracy and effectiveness of hate speech detection models on social media platforms.

Researchers have also focused on studying how the platform components/features affect the process of online hate detection. The user network which can be identified by analysing the following, followers, and fronds. and the user activities such as posting, interacting with retweets, favourites and likes shown to be related to the act of posting hate speech content. [32] prove that there is link between the high comment rate and the trolling. The more active a user is online, the more likely they are to engage in anti-social behavior. Additionally, researchers have identified more information about the content creator such as the gender and how it contributes in producing more or less hate [33]. Some studies found that there is a relation between directed hate or trolling and the Dark Tetrad of personality, such as trolling correlated positively with sadism, psychopathy, and Machiavellianism. Other studies also incorporate psychological features along with the textual features to enhance the online hate detection [34].

### *C. Incorporating actors of the online abuse event such as the target*

Recent studies have also emphasised the importance of incorporating the user contextual information to improve model performance [35]. They explored the integration of text and user-related context features, including the news article title, user screen name, and comments within the same thread. Their approach utilized both logistic regression and neural network models, resulting in a 3-4% improvement in F1 score compared to a strong baseline. Furthermore, combining these models led to an additional 7% increase in F1 score. This research underscores the significance of contextual information in accurately identifying subtle and creative language often employed in online hate speech. Building upon the importance of context, [36] proposed the Generalized Social Trend Model (GSTM) to measure and predict hate speech trends. Their approach incorporated various platform-related features, such as: geographical distribution, influential users, network nodedegree, Intense sentiment, exposure factors, temporal factors. The GSTM model provides an effective framework for analyzing hate speech dynamics across social media platforms. [36] analysis revealed notable differences in follower counts and language usage between users engaging in hateful speech and those producing counter-hate content. This comprehensive approach to hate speech trend prediction offers valuable insights into the complex nature of online hate speech propagation and its potential countermeasures. These studies collectively contribute to the growing body of research on context-aware hate speech detection and trend analysis, highlighting the multifaceted nature of online hate speech and the need for sophisticated modeling approaches to address this challenging problem.

In an adjacent area of research, there have been efforts tackling cyberbullying. For example, the comments' history of a user were used as a feature in in [37]. They also used users' characteristics and profile information. The results shows that user history of comments improves the cyberbullying detection accuracy compared to only analyzing individual comments. In addition, [38] show how a thread context improves the detection of cyberbullying. In this work, they mainly depend on the history of negative content and the related context of the platform which the model is based on. Cyberbullying is however different from other forms of abusive language such as hate speech, as cyberbullying tends to occurs in longer sessions and is recurrent [39], as opposed to shorter conversational exchanges, which is our focus here. A major shortcoming in current automatic hate speech detection research is the limited use of the target of online hate related contextual information. The primary focus has been on analysing the perpetrators or posts in isolation, without accounting for the role of the online hate targets and how incorporating such information can be a game changing. Incorporating target's available data could aid in accurately determining if a reply to a social median post should be classified as hateful or not in addition to the ability to classify whether the content that received abusive replies and /or content creator is considered to be hate prone or not.

## III. Methodology

In this section, we formulate our classification problem, describe the approach we take, and introduce the dataset and models we use for our research, as well as the feature engineering process.

### *A. Problem Formulation*

We define the conversational abusive language detection task as that where we aim to determine if a post that is replying to an earlier post is abusive or not. We define a conversation as a collection of replies R = $\{r_1,\ldots,r_i\}$ that are replying to a parent post, $p$. This forms a tree structure where each of the replies in R is directly linked to $p$, but the replies aren't linked to one another. For each reply $r_j$, we aim to determine the correct label in C = $\{abusive, non-abusive\}$.

The predictive function $f(\chi_i)$ is defined to minimize the predictive error of the predicted class label $y_i$ given the features $\chi_i$.

### *B. Approach*

We employ a supervised learning technique as the main text classification methodology, using our labelled dataset that contains replies annotated as abusive replies and non-abusive replies. This dataset is utilized to develop this classification task. This task provides predictions about the probability of a given reply being abusive using the above-mentioned features. Features such as parent tweet text content and tweet metadata are crucial for training models. During training, different combinations of these features are used as inputs for the models to effectively capture the correlation between the predictive features and the abusive replies.

Next, we describe the formulation of our classification experiment. Let $r_i$ represent a reply instance, which is represented with a set of features. To represent a reply vector, we use different permutations of the following feature families, hence investigating the impact and effectiveness of each feature family:

1) Text Content: The text content of the parent tweet in which this reply is directed to $\tau_i$ denoted as:
$Te_i = [e_{i1}, e_{i2}, \ldots, e_{in}]$
2) Parent-Tweet features: The parent tweet metadata features expressed in $\tau_i$ denoted as:
$Tw_i = [w_{i1}, w_{i2}, \ldots, w_{in}]$
3) Direct-reply-Tweet features: The direct reply tweet metadata features expressed in $\tau_i$ denoted as:

$Ru_i = [u_{i1}, u_{i2}, ..., u_{in}]$

4) Parent-Tweet Meta text features: Text metadata of the parent tweet features of tweet $\tau_i$ denoted as: $Mt_i = [m_{i1}, m_{i2}, ..., m_{in}]$
5) Direct-reply Meta text features: Text metadata of the parent tweet features of tweet $\tau_i$ denoted as: $Mr_i = [n_{i1}, n_{i2}, ..., n_{in}]$
6) Account features: Account of the parent tweet creator features, $\tau_i$ including all account related metadata features denoted as: $Ac_i = [c_{i1}, c_{i2}, ..., c_{in}]$

The classification prediction is mathematically represented as:

$$\hat{y}_i = f(\chi_i) = f([Te_i, Tw_i, Mt_i, Ac_i, Ru_i, Mr_i]) \quad (1)$$

The feature vector $\chi_i$ for reply instance $r_i$ is built with different permutations of the above features:

$$\chi_i = [Te_i, Tw_i, Mt_i, Ac_i, Ru_i, Mr_i] \quad (2)$$

### C. Dataset

As a dataset consisting of full conversations including replies to an initial parent post, we use the Online Abusive Attacks (OAA) dataset[1] [24]. This target-oriented dataset is specially designed to perform such experiments that captures all platform components. It comprises 2,371 distinct target accounts which are the accounts of the parent tweets creators and 106,914 conversations sparked by tweets posted by these accounts. A conversation refers to a parent tweet that has at least one reply from another user.[2] The dataset consists of 153,144 initial replies directed to the parent tweet. The labelling and annotation tasks were completed using Google Jigsaw's Perspective API [40], with manual validation of annotations showing reasonable agreement with the API's labels. In summary, the OAA dataset provides a valuable source of information for analysing and forecasting online abusive attacks, offering a detailed context and target-focused perspective. Table I provides the main statistics about the OAA dataset.

TABLE I
STATISTICS OF THE FINAL OAA DATASET AS USED IN OUR STUDY.

| Feature | Count |
|---|---|
| Number of user accounts | 2,367 |
| Number of conversations | 106,914 |
| Number of conversations with abusive replies | 21,383 |
| Number of conversations with non-abusive replies | 85,531 |
| Number of replies | 153,144 |
| Number of abusive replies | 24,907 |
| Number of non-abusive replies | 128,237 |

The dataset contains a holistic collection of conversations incorporating user and textual features, which we group into four types of features for our experiments, which we describe later.

### D. Classification Models

This section presents the models we use. The chosen models have different strengths and were selected based on the task requirements, dataset size, need for capturing context, and the trade-off between interpretability and performance.

These models are selected for their specific strengths in handling different types of data and tasks:

- Logistic Regression (LR): This model is chosen for its simplicity and ease of interpretation, making it ideal for understanding basic patterns in data, especially for binary classification tasks that can be adapted for multiclass classification.
- Support Vector Machine (SVM): SVM [41] is chosen for its effectiveness in high-dimensional spaces, which is beneficial for text classification tasks where the feature space can be very large. It is particularly good at finding the optimal hyperplane that separates different classes, making it suitable for tasks where the data is not linearly separable, e.g., through discriminative models.
- Random Forest (RF): Selected for its robustness against overfitting and ability to handle numerous features, Random Forest is an ensemble method effective for capturing complex data patterns by combining multiple decision trees.
- BERT model: The pre-trained transformer-based model, BERT [42], specifically 'bert-base-uncased', is selected for handling this classification task involving text data due to its bidirectional nature, which allows it to capture rich contextual information from both directions in the input text. The model's architecture enables it to understand complex relationships between words and their context. In this work, the BERT model was fine-tuned on the OAA dataset, adapting its pre-trained language understanding to the nuances of this classification task. The model's output is combined with additional meta-features layer, allowing it to leverage both textual and numerical information for more accurate predictions.

  Hence, the BERT model generates embeddings from the textual input, which are then concatenated with additional meta-features. As such, the BERT model needs a textual input that is then combined with other features, and therefore we limit BERT experiments to feature sets that include textual features and exclude feature sets without any text from our experimentation.

### E. Text Preprocessing

For the text classification models but excluding BERT, we perform a preprocessing step for textual input. We follow a text processing pipeline that consists of a sequence of steps that involves transforming raw text data into a structured format

[1] https://github.com/RaneemAlharthi/Online-Abusive-Attacks-OAA-Dataset [2]https://help.twitter.com/en/using-x/x-conversations

suitable for modeling. This pipeline consists of the following stages:

- Tokenization: This initial process is responsible for splitting the text into individual space-separated tokens.
- Stopword and Special Character Removal: We remove stopwords and special characters as less meaningful features in the classification process. We use the NLTK[2] (Natural Language Toolkit) and spaCy [3] libraries to achieve stopword removal. We then remove the following special characters: punctuation marks, symbols, and others that are not a word character or a whitespace character, etc., non-ASCII characters (including emojis, certain special characters, accented letters, and other symbols outside the standard ASCII range), extra spaces (including multiple consecutive spaces and leading and trailing spaces), Unicode numbers, single-letter words.
- Stemming: The third step involves performing a stemming process in order to reduce words to their base or root forms.

### *F. Context aware feature extraction and engineering*

Text features. This section explains all the steps we took for feature extraction and engineering. Starting by the extraction process for all the text related features including the parent tweet text and all its directed replies. The text preprocessing is different for the BERT model, and hence we define two separate text preprocessing methods next for the different types of models:

- LR, SVM and RF: We generate vectors with token counts, using both unigrams and bigrams. We tested both Bag of Words (BoW) and Term Frequency-Inverse Document Frequency (TF-IDF) initially; as the BoW approach led to better performance, we end up using it with the dimensionality restricted to 5,000 dimensions. In addition to token counts using BoW, we append features with sentiment scores for keywords matching a sentiment lexicon, providing positive or negative sentiment scores with additional information added to the vectors for lexicon keywords.
- BERT: we directly use the BERT embeddings generated by the model as the representation of the textual input.

Contextual features. In this experiment, we explore the effectiveness of different feature categories that reflect the context of the online abuse in the online conversational form. The conversations are composed of parent tweet as the main content generated by the target user, and a set of replies to that tweet. Each classification instance for us involves a single reply along with the parent tweet, and hence we derive features from this parent-reply pair. The features categorized as listed below.

1) Reply text (Rt): The reply text includes only the textual content of the replying post, overlooking all context from the conversation. We use this as the baseline feature set that we aim to compare the rest of the feature sets that do incorporate contextual information from the conversation for comparison.
2) Text features (Te): The text features include all text presented in the captured context of a complete conversation sample, which is the current reply and parent tweet that we are classifying at the moment.
3) Text meta features (Mt): It includes all additional information and attributes associated with the text without providing the exact text, such as stemmed character, hate word counts, negative word counts, positive word counts, abusive word counts, character count of parent tweet.
4) Tweet-based features (Tw): Tweet-based features are the features related to the tweet and the text of the tweet, such as hashtags, mentions, hate, abuse in the text content, etc.
5) Account-based features (Ac): Account-based features are the features that describe the user's account (the target's accounts only), such as follower count, favourite tweet count, etc. This group of features enables us to assess to which it is the user's characteristics that motivate others to post abusive replies to them, or it is instead the posts, as captured by the other three feature sets.

### *G. Training details*

All models used K-Fold Cross-Validation with 5 splits. SMOTE (Synthetic Minority Over-sampling Technique) applied to balance the training data. The text input was preprocessed using tokenization and padding to a maximum sequence length of 300.Meta features were standardized using StandardScaler.

For the BERT model: A pre-trained BERT model used as a first layer for the text encoder set to be trainable, for the finetuning. Additional input for meta-features, the BERT output is concatenated with the meta-features. Two dense layers were added with ReLU activation, each followed by dropout, and finally the output layer with a sigmoid activation. We run the model using a batch size of 32 and for 5 epochs.

### *H. Evaluation Metrics*

We report performance scores based on precision and recall, and the F1 score as the harmonic mean of precision and recall:

$$F1_score = 2 \times \frac{Precision \times Recall}{Precision + Recall} \tag{3}$$

While we report all three scores, our primary focus in on the F1 score, as we are interested in achieving a good balance of precision and recall.

To enhance the interpretability of our machine learning models and gain insights into feature importance, we report importance scores derived from a Random Forest model.

## IV. Results

Our experiments aim to look at how incorporating the target's information derived from the parent tweet as a complementary context can help with the detection of abusive content in replies. In what follows, we present the results of our experiments.

[2] https://www.nltk.org/

[3] https://realpython.com/natural-language-processing-spacy-python/

and that sole reliance of content from replies is insufficient. Feature combinations. Having seen that contextual features (rows 1-15) outperform the sole use of reply content (row 16), we are interested in further comparing the performance of combinations of different contextual features. We have tested combinations including only one feature type (rows 1-4), two feature types (rows 5-10), three feature types (rows 11-14) and all four feature types (row 15). Comparing these four different groups of results, we observe a general tendency for bigger combinations of features to lead to better performance.

| | Features | | | | | LR | | | SVM | | | RF | | | BERT | | |
|---|---|---|---|---|---|---|---|---|---|---|---|---|---|---|---|---|---|
| # | Rt | Te | Mt | Tw | Ac | F1 | Prec | Rec | F1 | Prec | Rec | F1 | Prec | Rec | F1 | Prec | Rec |
| 1 | | X | | | | 0.65 | 0.53 | 0.84 | 0.69 | 0.59 | 0.82 | 0.73 | 0.71 | 0.75 | 0.70 | 0.83 | 0.61 |
| 2 | | | X | | | 0.34 | 0.21 | 0.91 | 0.68 | 0.58 | 0.82 | 0.83 | 0.98 | 0.72 | – | – | – |
| 3 | | | | X | | 0.53 | 0.71 | 0.43 | 0.47 | 0.72 | 0.35 | **0.88** | 0.90 | 0.85 | – | – | – |
| 4 | | | | | X | 0.32 | 0.19 | 0.86 | 0.34 | 0.21 | 0.87 | 0.17 | 0.38 | 0.11 | – | – | – |
| 5 | | X | X | | | 0.71 | 0.63 | 0.81 | 0.73 | 0.68 | 0.79 | 0.81 | 0.90 | 0.73 | 0.74 | 0.89 | 0.63 |
| 6 | | X | | X | | 0.78 | 0.69 | 0.89 | 0.79 | 0.72 | 0.87 | 0.86 | 0.88 | 0.84 | 0.80 | 0.87 | 0.74 |
| 7 | | X | | | X | 0.69 | 0.60 | 0.83 | 0.72 | 0.65 | 0.81 | 0.75 | 0.81 | 0.70 | 0.70 | 0.79 | 0.63 |
| 8 | | | X | X | | <u>**0.91**</u> | 0.91 | 0.92 | 0.52 | 0.74 | 0.40 | 0.87 | 0.92 | 0.82 | – | – | – |
| 9 | | | X | | X | 0.35 | 0.22 | 0.84 | 0.75 | 0.73 | 0.78 | 0.83 | 0.98 | 0.72 | – | – | – |
| 10 | | | | X | X | 0.54 | 0.72 | 0.43 | 0.46 | 0.73 | 0.33 | 0.85 | 0.92 | 0.79 | – | – | – |
| 11 | | X | X | X | | 0.79 | 0.73 | 0.88 | 0.80 | 0.77 | 0.83 | 0.84 | 0.90 | 0.80 | **0.82** | 0.85 | 0.79 |
| 12 | | X | X | | X | 0.72 | 0.64 | 0.81 | 0.74 | 0.70 | 0.80 | 0.81 | 0.90 | 0.74 | 0.75 | 0.87 | 0.66 |
| 13 | | X | | X | X | 0.79 | 0.71 | 0.90 | **0.82** | 0.77 | 0.86 | 0.85 | 0.92 | 0.80 | 0.81 | 0.90 | 0.74 |
| 14 | | | X | X | X | 0.56 | 0.73 | 0.46 | 0.52 | 0.73 | 0.41 | 0.87 | 0.94 | 0.80 | – | – | – |
| 15 | | X | X | X | X | 0.81 | 0.74 | 0.89 | **0.82** | 0.80 | 0.84 | 0.85 | 0.90 | 0.81 | 0.80 | 0.85 | 0.75 |
| 16 | X | | | | | 0.74 | 0.67 | 0.85 | 0.75 | 0.68 | 0.84 | 0.84 | 0.96 | 0.74 | 0.70 | 0.84 | 0.59 |

TABLE II

THE MEAN OF F1, PRECISION, RECALL SCORES FOR THE 5-FOLD CROSS VALIDATION OF THE BINARY CLASSIFICATION TASK. THE HIGHEST SCORES IN EACH INDIVIDUAL MODEL (REPRESENTED BY BOLD TEXT) AND THE OVERALL HIGHEST VALUE ACROSS ALL MODELS (REPRESENTED BY BOTH BOLD AND UNDERLINED TEXT).

Table II presents the results of our experiments, showing results for four different models (LR, SVM, RF, BERT) and 16 different combinations of features; we refer to these combinations of features by the row number as indicated in the leftmost column of the table. Results for the BERT model are limited only to combinations of features that include at least a textual input, due to the dependency of the model on having some textual input which is the concatenated with other features, and as such combinations not including textual features were discarded.

Contextual vs non-contextual features. First, we look at the differences between contextual vs non-contextual features, to answer our primary research question about how leveraging conversational features including those derived from the parent tweet relating to the target can support the classification process. Hence, we compare the non-contextual model leveraging only reply content (row 16) with the remainder of contextual models (rows 1-15). We observe that, for all models, there are always combinations of contextual features which lead to improved performance over the non-contextual features, demonstrating that features derived from the parent are useful

With exceptions, such as in the case of RF, we observe that using a single feature type (rows 1-4) leads to substantially lower F1 scores, often in the range between 0.3 and 0.7. Performance gradually improves as more feature types are incorporated, with better performances when 2-4 feature types are incorporated.

There are exceptions. The RF model is surprisingly consistent and can perform reasonably well with a single feature type already. While the LR model shows a general tendency to improve when using more features, its overall best performance is achieved when using two feature types combining Mt and Tw. Overall, however, results show that it is a safer choice to rely on more feature types, as in those cases models are less likely to underperform as it can happen when using fewer feature types.

Feature types. While we see that combining more feature types is generally a safer choice, do all features contribute the same and should we incorporate them all? And what does the effectiveness of each of the features tell us about the contextualized abusive language detection task?

Our results suggest that the content-based feature types (i.e. Te, Mt, Tw) are the ones contributing the most to the performance improvement. For example, the combination of these three feature types (Te, Mt, Tw) performs well across all models, and performs almost as well as the combination of all four feature types (Te, Mt, Tw, Ac). The fact that removing the Ac features leads to almost no performance loss indicates that the three first features suffice and that account-based features (Ac) contribute little to nothing.

This finding is further reinforced when we look at the combination using only Ac features (row 4). This combination is consistently poor across all models, with performance scores in the range between 0.17 and 0.34. Hence, we can conclude that account-based features do not help with the classification task and it is primarily the content-based features that do. This in turn suggests that account-based features of the target of a post are not indicate of a reply being abusive, that it is best to rely on content only for the classification.

## V. Results: Further delving into the features

So far we have look at the overall F1 scores, and how different features contribute to the overall performance. However, our dataset contains multiple different target users to whom the replies are directed. Does the classification performance vary across different target users? Is the performance similar across all target users?

To look at this, we break down the performances by groups of target users, to see how performances differ. We make two groups, target users for whom performance scores are best, compared to target users for whom performances are scores are lowest. Looking at each target user individually, we can calculate the F1 score of our model for each target user. Having this, we calculate the median F1 scores of our prediction performance across all users. Having this median, we identified the 50% target users whose performance is above the median (above-median), and the 50% target users whose performance is below the median (below-median). We next analyze features of above-median vs below-median users next to identify what leads to improved performance.

### *A. Analysis of features for above-median and below-median users*

Meta-text based features. For the meta-text based features in Figure 2 we started by identifying highly predictive features based on significant differences in average values between the high-performance above-median and low performance belowmedian groups. The following features including: Parent Word Count, Parent Character Count, Parent Sentence Count, Parent Average Word Length, Parent Hashtag Count, Parent URL Count, Parent Punctuation Count, Parent Average Sentence

Length, DirectReply Average Word Length, and DirectReply Average Sentence Length, with higher averages in the abovemedian group of the previously mentioned features, these features demonstrate to be strongly associated with better performance.

Conversely, features like: DirectReply Sentence Count, DirectReply Stopword Count, and DirectReply Capitalized

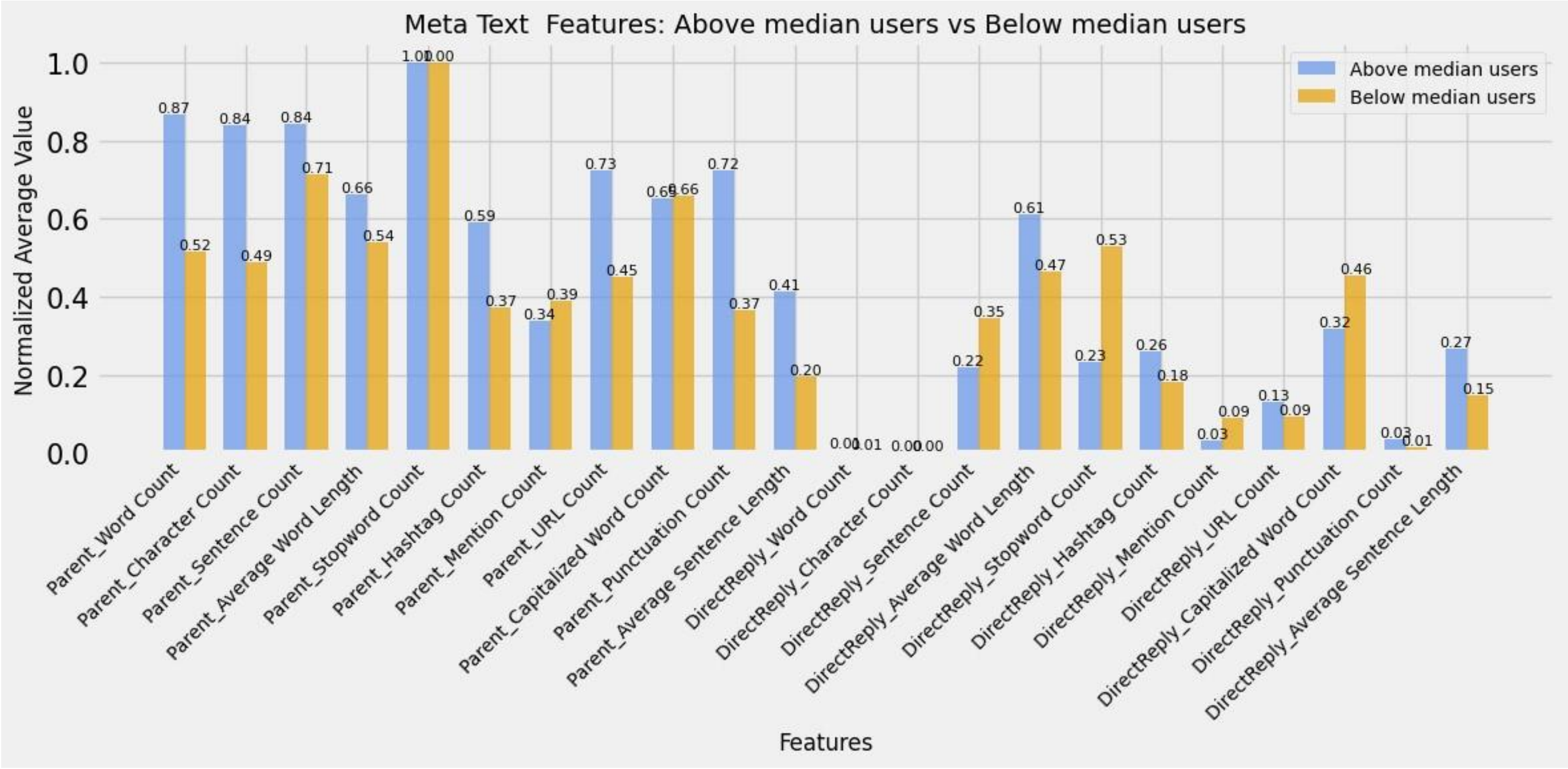

Fig. 2. A Comparison of normalized average feature values for Above median users (blue) and Below median users (orange)for the meta-text based features

Word Count, are associated with higher averages in the belowmedian group, indicating poorer performance.

The following features, including: word, character, and sentence count, hashtag, URL, Punctuation count of the parent tweet alongside the average word and sentence length for both parent and direct reply. They exhibit significantly higher average values for the above-median group compared to the below-median group, hence suggesting that higher values of such features ranging between 0.87 and 0.59 are strongly linked to a better model performance.

Features related to the direct reply such as: DirectReply Sentence Count, DirectReply Stopword Count, DirectReply Capitalized Word Count, show significantly higher average values for the below-median group, implying that higher values correlate with poorer performance, while lower values are associated with better outcomes which range from 0.22 to 0.32.

On the other hand, Parent Stopword Count, Parent Mention Count, Parent Capitalized Word Count, DirectReply Word Count, DirectReply Character Count, DirectReply Hashtag Count, DirectReply Mention Count, DirectReply URL Count, DirectReply Punctuation Count features demonstrate no significant differences in average values between the groups, indicating minimal predictive power for distinguishing high versus low performance target users. This suggests that they have a limited impact on model performance.

Tweet-based features. Figure 3 showing the averages for above-median and below-median users for tweet-based features shows overall marginal differences between averages. Both parent tweet number of retweets and favourites have slightly higher averages with 0.26 were associated with the below median users. The direct reply negative sentiment score averages of the below- and above-median users were equally distributed. For the direct reply positive sentiment score higher average with 0.34 where associated with the below median users. The neutral sentiment score isn't contributing significantly, while the name entity count high average of 0.13 differently associated with the above median users. Overall, tweetbased features show a marginal impact on model performance when we look at the two groups.

Account-based features. Figure 4 shows the normalized average values of the account-based features for above-median and below-median users.

We start with a general identification of the more predictive features based on the average value difference between abovemedian and below-median users. With the exception of some of the features, we observe that most of the account-based features have small differences between above-median and belowmedian users, again reinforcing the fact that account-based features are not as helpful for the prediction as the contentbased features are. Some of the features, such as: friends count, listed count, geo enabled, verified, statuses count, contributors enabled, is translator, default profile, default profile image, following, follow request sent and notifications exhibit some degree of difference between above-median and belowmedian users, while those with minimal differences including followers count, favourites count, is translation enabled, and has extended profile showed limited discriminatory power.

Despite the modest average differences for some of the account-based features, these are not substantial and are not consistent across the features. Compared to the greater differences we observed for the meta text features above, this reinforces the results of our experiments suggesting that account-based features make a marginal contribution to model performance.

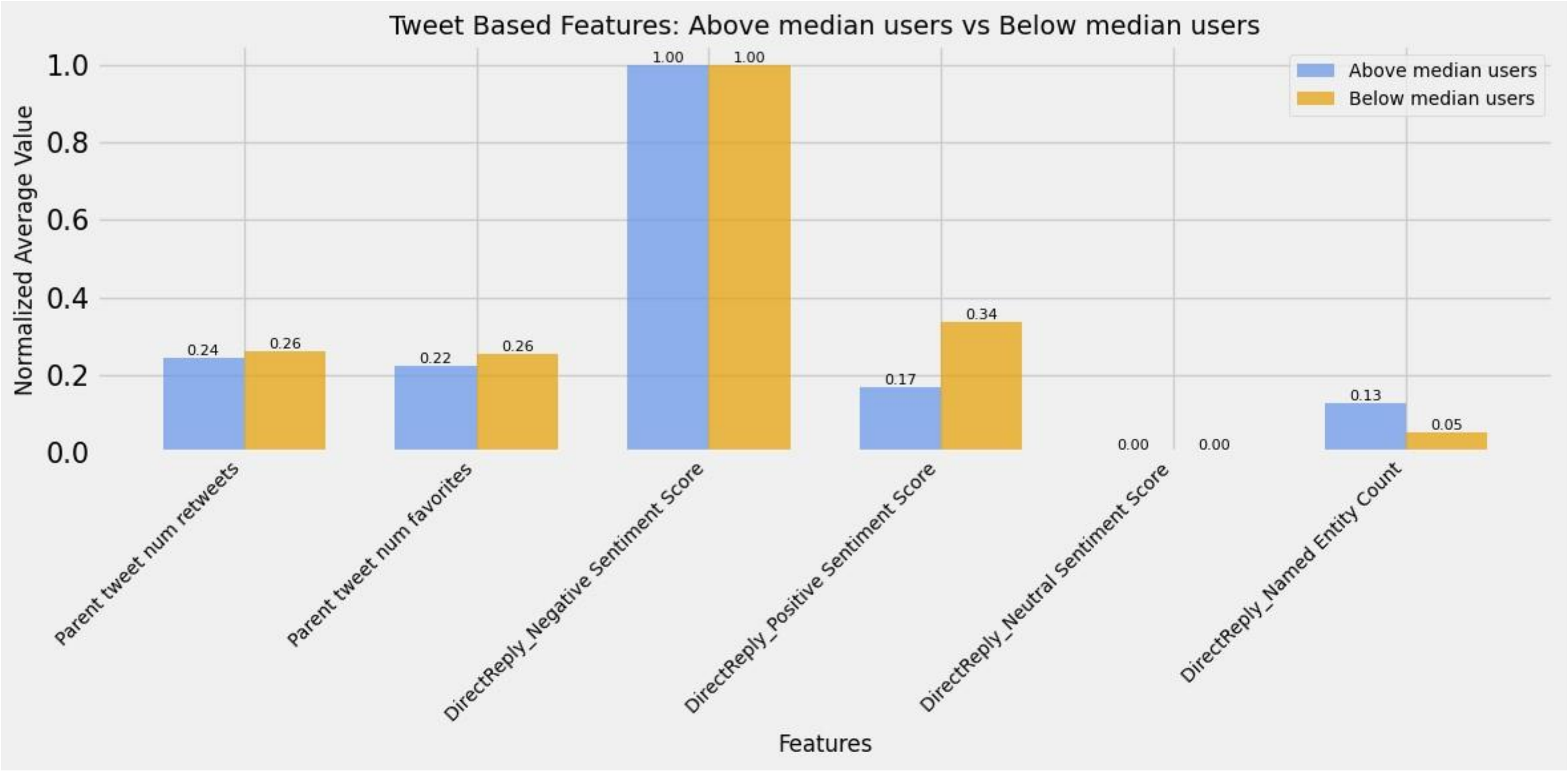

Fig. 3. A Comparison of normalized average feature values for Above median users (blue) and Below median users (orange) for the tweet based features

*B. Analysis of feature importance*

To further analyze the importance of each feature in the predictions, we perform a feature importance analysis derived from a Random Forest model, which allows quantitatively measuring the importance of each feature towards the predictions. We next look at the three groups of features, metatext based features, tweet-based features and account-based features.

Meta-text based features. Looking the importance scores of meta-text based features, as shown in Figure 5, we observe that the direct reply character count, average sentence length, stop word count, and word count shown to have the highest importance values. These results are surprising as one would not expect the length of the posts to be predictive of abusive language necessarily, but it may have to do with the content being more substantial and hence more prone to receive certain kinds of replies.

These features with the highest importance scores are followed by the parent tweet related meta-text features such as the parent average sentence length, and word length, and parent word count. After that, the direct reply average word length and the punctuation count shown to be less important features which means that it has a relatively minor impact on the model's predictions. On the other hand, features related to the embedded URLs, hashtags, mention counts for both direct replies and parent tweets identified as features with the lowest importance scores along with the sentence count and the capitalized word count. It is important to note that the stop word count of the direct reply considered to be among the top three high important features while the stop word count of the parent tweet is less important.

Tweet-based features. In Figure 6 we show the importance scores for tweet-based features. We see that the direct reply negative sentiment score is the most important feature among the tweet-based features. This is followed by the direct reply neutral and positive sentiment score, with a lesser importance for the parent tweet number of retweets and favourites. Finally the direct reply named entity shown to have the lowest importance score.

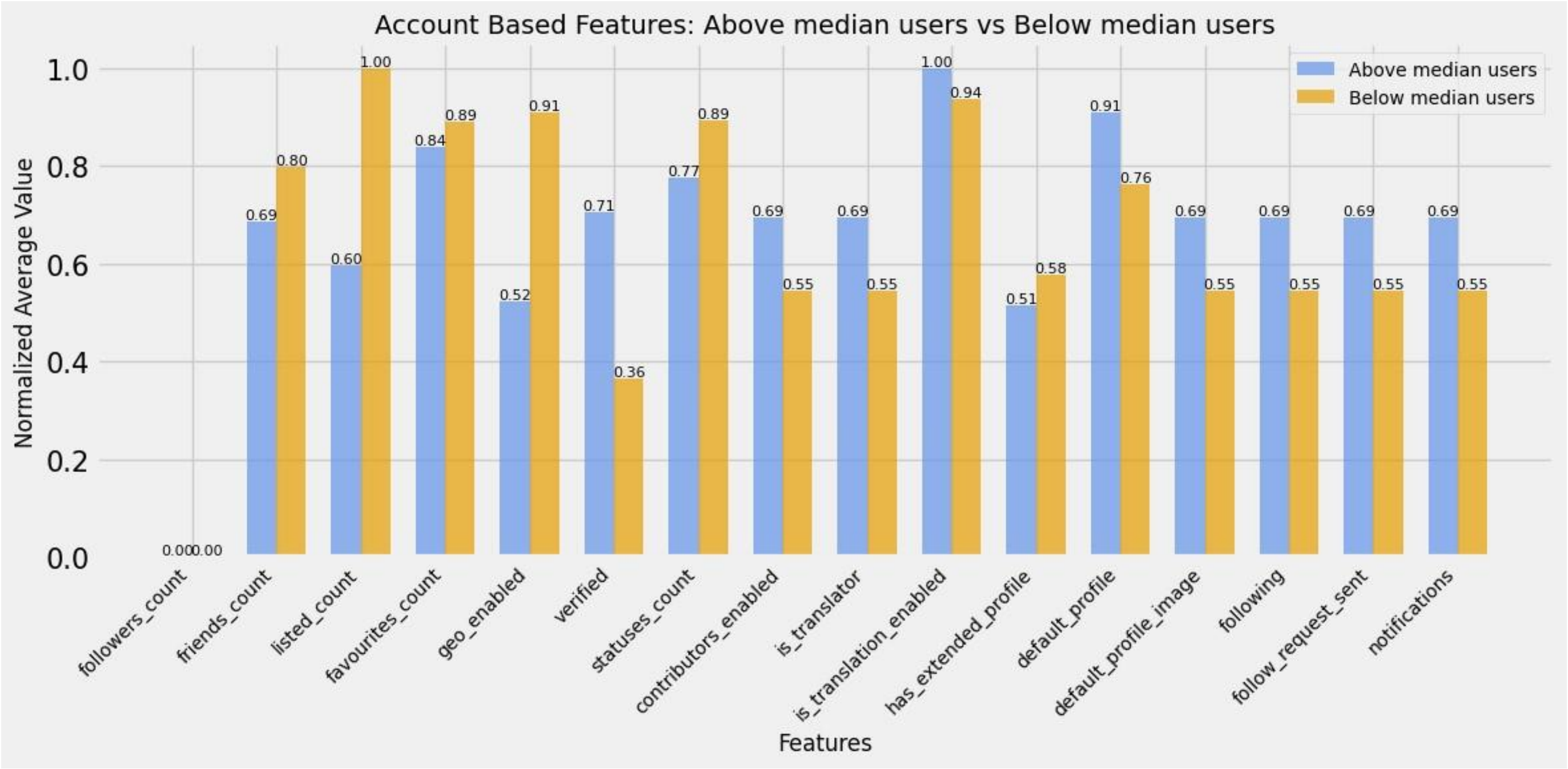

Fig. 4. A Comparison of normalized average feature values for Above median users (blue) and Below median users (orange) for the account based features

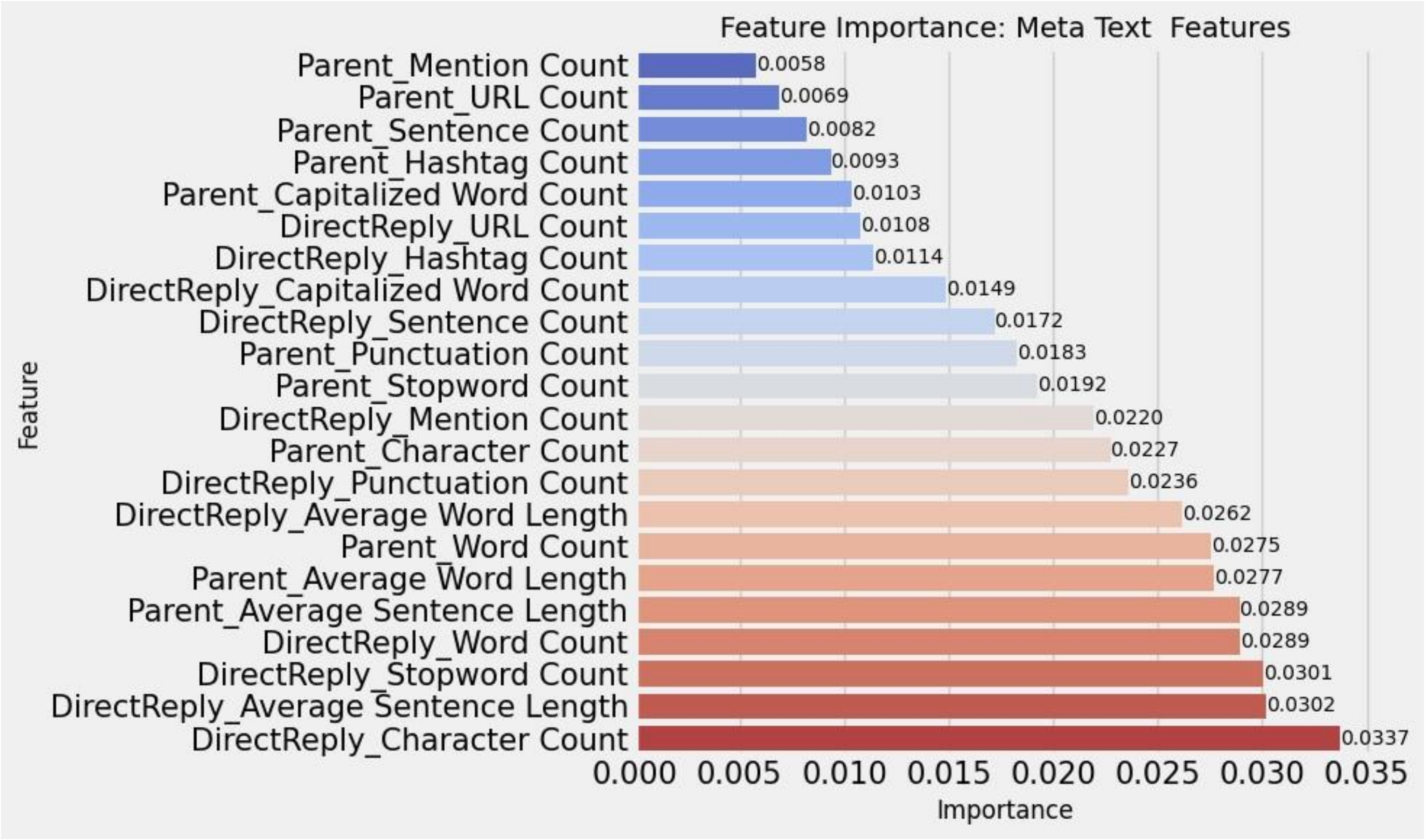

Fig. 5. Features importance for the meta based features.

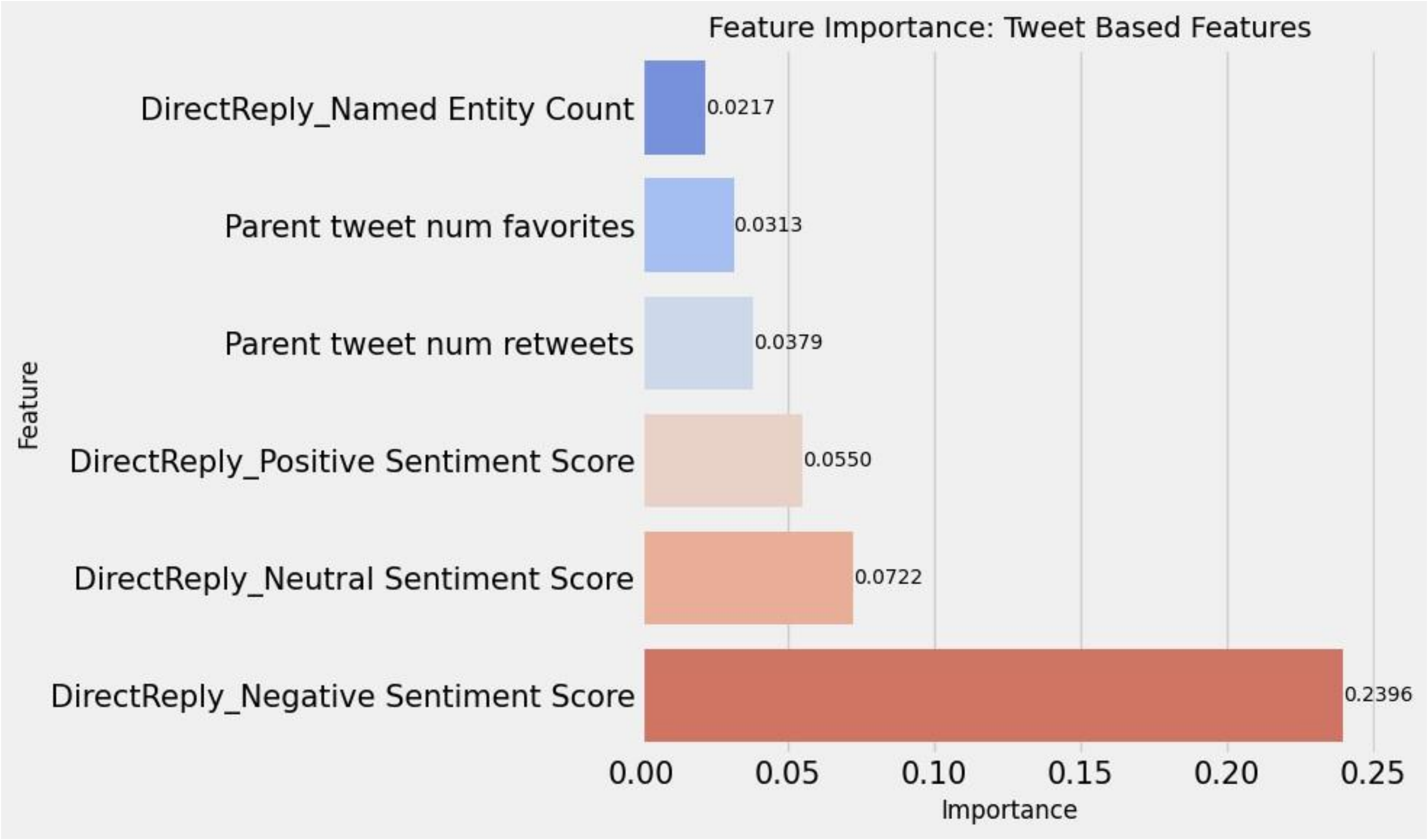

Fig. 6. Feature importance for the tweet based features.

Account-based features. In Figure 7 we show the importance scores for account-based features. Among these features, we see that the favourite count is the most important feature followed by the followers and friends counts. These features reflect the level of popularity and engagement that the target user attracts, hence suggesting that these users are more likely to attract abusive replies; however, they are unlikely to provide enough predictive support as observed in the lack of positive impact in our experiments.

The verification status of the account surprisingly shows a very low importance score, however this is likely because only a small number of users are verified. The same applies to the geo enabled status feature, which only has a positive value for a small number of users. Other features are less important.

## VI. Discussion: Revisiting the Research Questions

This section provides a discussion on the experiment findings and how these findings can answer the main research questions.

- RQ1: How accurately can we predict if a reply to a tweet is abusive or not based on the target's related features as a complementary context information of the direct reply?

Our experiments demonstrate the importance of leveraging contextual information in conversational settings to determine if a reply is abusive or not. In our experiments, we have looked at a large collection of conversations across different targets, and studied how the use of contextual features derived from both the reply and the parent tweet compared to the widely studied approach in the literature of solely relying on the (reply) tweet's content itself. Our study finds that context can substantially boost performance in abusive language detection, showing that the non-contextual approach always underperforms. Among the contextual approaches, we observe some variation across different classification models, but in general they show a tendency towards variants using more features to perform best.

- RQ2: What categories of features are able to predict solely and enhance the prediction when it's combined with other features?

Through our experiments, we observe that greater combinations of features tend to lead to better performance. Where we have studied four different families of features, only using a single feature family tends to underperform, with combinations of 2, 3 or 4 feature families performing typically better. Among the feature types, we observe that account-based features are the least useful ones; in fact, if we simply use account-based features, we observe very low performances suggesting that these features are not helpful for the prediction. This is further reaffirmed with the combinations of features, where we observe that com-binations of features incorporating account-based features do not improve performance over the same combination excluding account-based features. On the positive side, we observe that it is content-based features, specifically meta-text and tweet-based features, that have a positive impact on model performance. The latter are in fact the features that most contribute to model performance and which are the ones that are safest to use, suggesting that, for abusive language detection in conversational settings, it is best to rely on content derived from the context, but not on the authors.

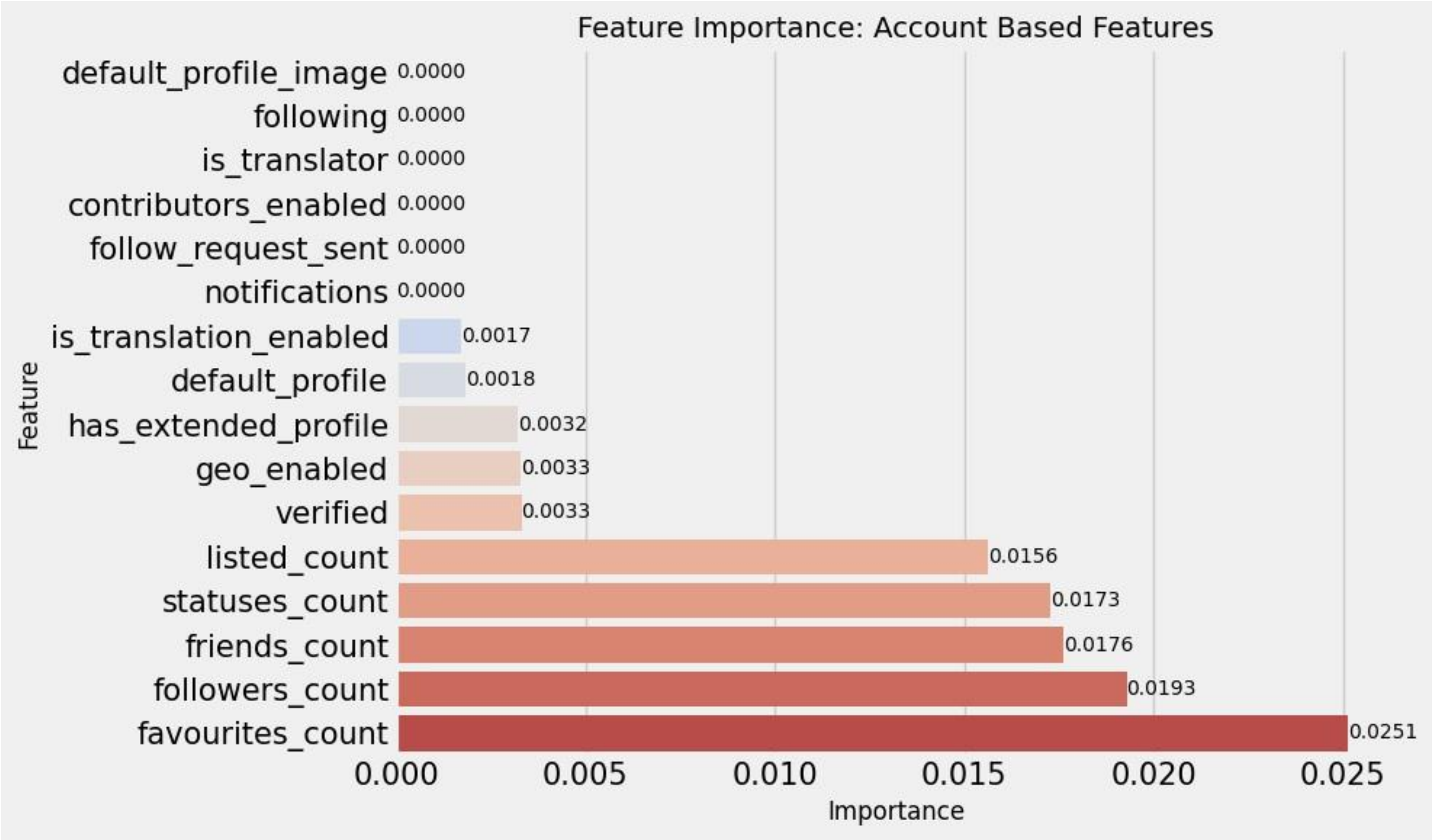

Fig. 7. Features importance for the account based features.

## VII. Conclusion

Our study investigates the ability to predict if a social media reply to a previous post is abusive or not in a conversational setting. This enables us to study contextual features derived from the conversation, assessing the extent to which context can help with the task as well as to study the types of features that contribute to this classification.

Using four different classification models on a dataset of conversational exchanges where replying posts are labelled as abusive or not, we perform experiments studying the impact of different features. We find that the traditional approach of simply using a social media post's own content to determine if it is abusive can be quite limited, and that this model can be substantially improved by leveraging contextual features derived from the conversation. Among the types of features that one can exploit from the context of the conversation, we find that content-based features are the ones that contribute positively to the prediction task, whereas account-based indicating who the target is, are not useful. All in all, this suggests that, for abusive language detection, one should aim to leverage surrounding context, but this should focus on content rather than who the users are. Focusing on contentbased features, we observe that to achieve competitive results it is a safer choice to rely on greater combinations of more feature types, as these combinations tend to lead to improved performance. We also perform a deeper study into individual features, which provides insights into how each of the features can contribute to the task.

While our research advances research in contextualized abusive language detection in conversational settings, it is not without limitations. Our research is limited to data in the English language, and future research could look into other languages to look into the generalizability of findings across languages. Moreover, our study of features has been limited to those features available to us; ideally, one may also want to look at additional features, for example features derived from the social networks of users (e.g. who they follow and who they are followed by), who users interact with, etc.